\documentclass[11pt]{article}

\usepackage[utf8]{inputenc}
\usepackage[T1]{fontenc}
\usepackage[english]{babel}
\usepackage{mathptmx}          % Times text + math (widely available on arXiv)
\usepackage[scaled=0.9]{helvet}
\usepackage[letterpaper,margin=1in]{geometry}
\usepackage{graphicx}
\graphicspath{{../docs/evaluation/}{../docs/training/}{./figures/}{./}}
\usepackage{booktabs}
\usepackage{amsmath}
\usepackage{xcolor}
\usepackage{microtype}
\usepackage{caption}
\usepackage[hidelinks]{hyperref}
\usepackage{url}
\usepackage{authblk}

\title{\bfseries Manac\'a-1B: An Open, Reproducible Brazilian-Portuguese\\
Language Model and a Tokenizer-Aware, Paired Evaluation}

\author[1]{Bruno Leonardo Santos Menezes}
\author[1]{Carlos Leonardo Souza Cardoso}
\author[1]{Fabio Andre Machado Porto}
\affil[1]{Laborat\'orio Nacional de Computa\c{c}\~ao Cient\'ifica (LNCC),
Petr\'opolis, Brazil\\ \texttt{\{brunolsm, cardoso, fporto\}@lncc.br}}

\date{Preprint. \today}

\begin{document}
\maketitle

\begin{abstract}
\noindent
Brazilian Portuguese remains under-served by open language models, and the few
that exist are difficult to reproduce and are often compared without measures of
uncertainty. We release Manac\'a-1B, an open decoder-only model of 1.72 billion
parameters trained from scratch for Brazilian Portuguese with a fully
containerized, reproducible pipeline. The pretraining is stable, with zero skipped
or NaN steps and self-recovering loss spikes, and we release its full log and
dynamics. We evaluate the model against nine open baselines on four Portuguese
benchmarks under a single harness. Every comparison
reports a standard error and a paired significance test, and the harness is
validated against previously published numbers. On last-word prediction Manac\'a-1B
is the strongest model below the 7B scale, exceeding both Tucano-1b1 and Tucano-2b4
on LAMBADA-PT with large paired margins; it is competitive on commonsense
completion and near chance on multiple-choice reasoning, as are all small base
models. Along the way we document a concrete evaluation pitfall: converting a
SentencePiece tokenizer with case-folding normalization to the HuggingFace fast
format silently drops the normalizer, routing every capitalized token to
byte-fallback and depressing scores in a way that is invisible in aggregate
metrics. The uncorrected tokenizer lowered LAMBADA-PT accuracy from 45.3 to 25.0;
we quantify the effect and provide a one-line fix that reproduces the training
tokenizer exactly. Code, raw training and evaluation logs, per-example prediction
vectors, the model weights, and the corrected tokenizer are released so that every
number in this paper can be recomputed.
\end{abstract}

\section{Introduction}

Open language models for Brazilian Portuguese have grown in number but not yet in
reproducibility or in the rigor of their evaluation. Tucano \cite{tucano} and its
TeenyTinyLlama predecessor \cite{ttl} released model families and the GigaVerbo
corpus; Sabi\'a adapted LLaMA to Portuguese at the 7B scale \cite{sabia}; Gl\'orIA
trained a European-Portuguese decoder and introduced the CALAME-PT last-word
benchmark \cite{gloria}. These works advanced the field, yet three gaps persist.
The training pipelines are rarely reproducible end to end, comparisons between
models are usually reported as point estimates without uncertainty, and the
tokenizer, which silently governs how each model reads text, is seldom audited.

We address these gaps with a model and a protocol rather than with scale. Manac\'a-1B
is a 1.72B-parameter Brazilian-Portuguese model trained from scratch inside a
containerized Megatron-LM pipeline on a curated, openly licensed corpus of about 20
billion Brazilian-Portuguese tokens, with every configuration, log, and artifact
committed to a public repository. We evaluate it against nine open baselines that
span three tokenizer families and two orders of magnitude in size, on four
Portuguese benchmarks, under one harness, and we attach a standard error and a
paired test to every comparison. We also treat the tokenizer as a first-class
object of study, and we show that a common conversion step can invalidate an
evaluation without leaving a visible trace.

Our contributions are the following.
\begin{itemize}
  \item \textbf{An open, reproducible model.} Manac\'a-1B, trained from scratch for
  Brazilian Portuguese, with a fully containerized pipeline and with all code, raw
  training and evaluation logs, and per-example prediction vectors released.
  \item \textbf{A rigorous comparison.} Manac\'a-1B against nine open baselines on
  CALAME-PT, ARC-Challenge-PT, HellaSwag-PT, and LAMBADA-PT, under a single harness,
  with binomial standard errors, bootstrap confidence intervals, and paired McNemar
  tests, validated against published numbers.
  \item \textbf{A tokenizer-fidelity finding.} We show that HuggingFace conversion
  drops the SentencePiece case-folding normalizer, quantify the resulting invisible
  penalty, and release a corrected tokenizer that reproduces the training
  tokenization exactly.
\end{itemize}

\section{Model and Training Pipeline}
\label{sec:train}

Manac\'a-1B follows a modern decoder-only design and is trained by a pipeline whose
every step runs inside version-pinned containers. The architecture is a Llama-style
transformer with the parameters in Table~\ref{tab:arch}: 24 layers, a model
dimension of 2048, a SwiGLU feed-forward width of 8192, 32 attention heads with 8
key-value groups (grouped-query attention), rotary position embeddings, RMSNorm,
and, unlike many recent Llama variants, learned biases in all linear layers, which
we preserve through every conversion. The tokenizer is a SentencePiece model
\cite{sentencepiece} with a 64k vocabulary (padded to 64{,}128) and \texttt{nmt\_nfkc\_cf} normalization, which
applies NFKC and case folding; the model is therefore lowercase by construction, a
design choice inherited from the LLM-jp recipe that we discuss in
Section~\ref{sec:tok}.

Training used the LLM-jp fork of Megatron-LM \cite{megatron} with a distributed
optimizer, z-loss,
and bfloat16, for 20{,}000 optimizer steps at a global batch size of 512 and a
sequence length of 4096, which corresponds to about 41.9 billion tokens, or
roughly 24 tokens per parameter. This budget sits close to the compute-optimal
ratio of about 20 tokens per parameter \cite{chinchilla}, so Manac\'a-1B is a
near-compute-optimal model rather than an over-trained one. The Tucano family, by
contrast, is trained on hundreds of billions of tokens, more than an order of
magnitude beyond the compute-optimal point, and that difference in token budget,
not architecture, is the lens through which we read the results. The run was
stable: across all 2{,}000 logged iterations the counters report zero skipped and
zero NaN steps, the training loss decreases monotonically from 11.41 at the first
logged step to 2.48 at the last, and the gradient norm stays near 0.10 after two
early transients that recover without a skip. The raw training log is released with
the repository.

The optimization and systems configuration is chosen to reproduce the LLM-jp-3.1-1.8B
recipe on modest hardware rather than to chase a novel setup, and every value is
recorded twice, once in a per-run provenance file that captures the repository commit,
the effective hyperparameters, and the container versions, and once in the raw log we
release. Optimization uses Adam with $\beta_1=0.9$, $\beta_2=0.999$, and
$\epsilon=10^{-8}$, a weight decay of 0.1, and gradient clipping at a global norm of
1.0. The learning rate warms up linearly over the first 2{,}000 steps to
$3\times10^{-4}$ and then follows a cosine decay to $3\times10^{-5}$ over the full
20{,}000 steps, and a PaLM-style z-loss regularizes the output logits for numerical
stability. Weights are initialized with standard deviation 0.02 and a fixed seed of
1234. The run fits on two 24\,GB GPUs connected over PCIe without NVLink: because
tensor parallelism would be bandwidth-bound on that interconnect, we use pure data
parallelism (two-way) with a ZeRO-1 distributed optimizer and overlapped gradient
reduction and parameter gathering, a micro-batch of one with gradient accumulation to
the global batch of 512 (about 2.1 million tokens), full activation recomputation to
fit the 24\,GB budget, FlashAttention, and bfloat16 throughout. Table~\ref{tab:opt}
lists the settings. The point of reporting them in full is that the entire run is a
single, reproducible command over version-pinned containers, not a bespoke cluster
job.

\begin{table}[t]
\centering
\caption{Architecture and training summary of Manac\'a-1B.}
\label{tab:arch}
\small
\begin{tabular}{ll}
\toprule
Parameters & 1{,}722{,}951{,}680 ($\approx$1.72B) \\
Layers / model dim / FFN dim & 24 / 2048 / 8192 (SwiGLU) \\
Heads / KV groups / head dim & 32 / 8 (GQA) / 64 \\
Position / norm / bias & RoPE ($\theta{=}5{\times}10^{5}$) / RMSNorm / all linears \\
Tokenizer & SentencePiece, 64k (\texttt{nmt\_nfkc\_cf}) \\
Context length / precision & 4096 / bfloat16 \\
Framework & Megatron-LM (LLM-jp fork), distributed optimizer \\
Steps / global batch / tokens & 20{,}000 / 512 / $\approx$41.9B ($\approx$24 tok/param) \\
Stability & 0 skipped, 0 NaN; loss 11.41 $\rightarrow$ 2.48 \\
\bottomrule
\end{tabular}
\end{table}

\begin{table}[t]
\centering
\caption{Optimization and systems configuration. All values are echoed in the released
provenance record and raw log.}
\label{tab:opt}
\small
\begin{tabular}{ll}
\toprule
Optimizer & Adam ($\beta_1{=}0.9$, $\beta_2{=}0.999$, $\epsilon{=}10^{-8}$) \\
Weight decay / grad clip & 0.1 / 1.0 (global norm) \\
Peak / min learning rate & $3\times10^{-4}$ / $3\times10^{-5}$ \\
Schedule / warmup & cosine decay / 2{,}000 steps (linear) \\
Regularization & z-loss (PaLM/LLM-jp style) \\
Init.\ std.\ / seed & 0.02 / 1234 \\
Global batch / micro-batch & 512 ($\approx$2.1M tokens) / 1 + accumulation \\
Parallelism & DP=2, TP=1, PP=1; ZeRO-1 distributed optimizer \\
Memory / attention & full activation recompute; FlashAttention \\
Hardware & 2$\times$24\,GB GPU, PCIe (no NVLink) \\
\bottomrule
\end{tabular}
\end{table}

\section{Pretraining Data}
\label{sec:data}

Manac\'a-1B is trained on a deliberately small, fully curated, and openly licensed
Brazilian-Portuguese corpus rather than on a large raw web crawl, a choice that trades
scale for provenance and reproducibility. The corpus draws on three sources, each under
an open license: a curated subsample of GigaVerbo \cite{tucano}, the largest open
Brazilian-Portuguese web corpus, which supplies most of the general-domain text; the
Ulysses Tesem\~o collection of Brazilian legal and legislative documents, which is in
the public domain; and the Portuguese Wikipedia. Table~\ref{tab:corpus} reports the
measured composition. After cleaning and within-source deduplication the corpus holds
about 20.1 billion tokens across 33.7 million documents, dominated by general web text
(73 percent) with a substantial legal component (24 percent) and an encyclopedic tail
(3 percent). It is entirely Brazilian Portuguese and contains no code or English, which
keeps the released model a clean single-variety base.

\begin{table}[t]
\centering
\caption{Measured composition of the Manac\'a-1B pretraining corpus, from the corpus
validation pass over 674 shards; token counts are estimated at about four characters
per token.}
\label{tab:corpus}
\small
\begin{tabular}{llrl}
\toprule
Source & Domain & Tokens & License \\
\midrule
GigaVerbo (subsample) & General web & $\approx$14.7B (73.1\%) & Apache-2.0 \\
Ulysses Tesem\~o & Legal, legislative & $\approx$4.8B (23.9\%) & Public domain \\
Wikipedia-pt & Encyclopedic & $\approx$0.6B (3.1\%) & CC BY-SA 4.0 \\
\midrule
Total & 33.7M documents & $\approx$20.1B & open \\
\bottomrule
\end{tabular}
\end{table}

Each source passes a light, source-appropriate cleaning pipeline rather than one
aggressive global filter. GigaVerbo, already curated by its authors, receives only
sanity filters on alphabetic ratio and length; the legal collection uses a lower
alphabetic threshold suited to statute text together with a domain-specific cleaner;
Wikipedia is cleaned and its article titles are prefixed to their bodies. Within-source
near-duplicate removal uses MinHash locality-sensitive hashing. Because the three
sources are already disjoint in domain and individually curated, we skip the
cross-source global deduplication that the pipeline reserves for noisier web sources,
and we record that decision rather than leaving it implicit. Every cleaned shard is
stored with its source, a per-document language tag, and a quality score, so the
composition of the training mix stays auditable after the fact.

The token budget of Section~\ref{sec:train} is best read against this corpus size. About
20.1 billion unique tokens trained for roughly 42 billion update tokens is close to two
epochs rather than a single pass, and repetition at this level is known to be nearly as
effective as fresh data up to about four epochs \cite{muennighoff}. The
near-compute-optimal ratio is therefore a statement about update tokens: each token is
seen about twice, which sits well inside the benign-repetition regime and is the honest
way to read the budget. Documents are concatenated in a fixed source order, marked with
end-of-document boundaries, and then shuffled by the framework under a fixed seed; there
is no curriculum, which we leave to future work.

\section{Training Dynamics}
\label{sec:dyn}

\begin{figure}[t]
\centering
\includegraphics[width=\textwidth]{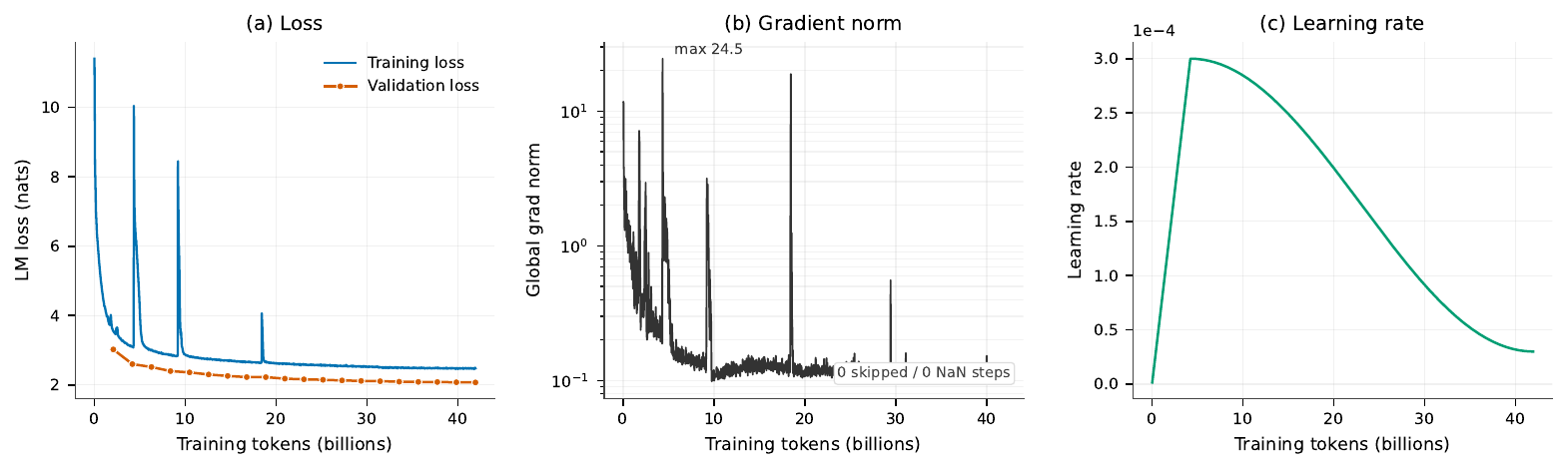}
\caption{Pretraining dynamics of Manac\'a-1B over roughly 42 billion tokens.
(a) Training loss and held-out validation loss; both fall and plateau, and the
validation loss is still declining at the end. (b) Global gradient norm on a log
scale; the run stays near 0.10 with a few self-recovering transients (largest 24.5)
and zero skipped or NaN steps. (c) Learning rate, a warmup followed by a cosine decay
to $3\times10^{-5}$. Regenerated from the raw log by
\texttt{scripts/eval/plot\_training\_dynamics.py}.}
\label{fig:dyn}
\end{figure}

The run was stable from scratch, and its behavior is worth showing in full because it
is one of the clearer signals that the pipeline is sound. Figure~\ref{fig:dyn}
summarizes the entire pretraining. Across all 2{,}000 logged iterations the framework
counters report zero skipped and zero NaN steps, so no update was ever discarded. The
learning rate follows a short warmup to a peak of $3\times10^{-4}$ and then a cosine
decay to $3\times10^{-5}$, and the schedule completes exactly as planned.

Both loss curves fall and then plateau, and the validation curve is the more
informative of the two. The per-iteration training loss drops from 11.41 to 2.48; the
validation loss, measured every 1{,}000 steps on a held-out split, decreases in step
with it and reaches 2.07 nats, a perplexity of 7.96, while still declining at the end
of the schedule. The validation loss sits below the noisier per-iteration training
estimate throughout, and it shows no upward turn, which indicates that the model does
not overfit and that, even at a near-compute-optimal budget, a larger token allotment
would likely continue to lower the loss, consistent with the training regime described
in Section~\ref{sec:train}.

The gradient norm tells the most interesting part of the story. For most of training
it stays near 0.10, but the run passes through a handful of sharp transients, the
largest reaching a global norm of 24.5, each of which coincides with a brief spike in
the training loss. Every one of these transients recovers within a step or two,
without a skipped update and without producing a NaN, and the loss returns to its
trajectory. Loss spikes of this kind are common in from-scratch language-model
training; the point here is that they were absorbed by the optimizer and the z-loss
regularization rather than destabilizing the run, which is what a healthy, reproducible
pretraining looks like.

\section{The Tokenizer as an Object of Study}
\label{sec:tok}

The tokenizer is a SentencePiece unigram model trained on a Portuguese sample of the
corpus with a vocabulary of 64{,}000, a character coverage of 0.9995, byte fallback
enabled so that no input is ever out of vocabulary, digit splitting, and the
\texttt{nmt\_nfkc\_cf} normalization rule, which applies NFKC and case folding before
segmentation. Four control tokens (unknown, beginning, end, padding) and a small set
of reserved chat-template symbols are declared explicitly. When Megatron pads the
vocabulary to a multiple of 128 the working size becomes 64{,}128; the extra rows are
padding that is never emitted, and the evaluation code masks them.

A lowercase model is only evaluated fairly if the text reaches it exactly as it did
in training, and this is easy to get wrong. Because \texttt{nmt\_nfkc\_cf} folds case,
the model is lowercase by construction. When we converted the model to the HuggingFace
format for evaluation with community harnesses, the resulting fast tokenizer carried
no normalizer at all. As a consequence, any capitalized character fell back to
byte-level tokens that never appear in training, while lowercase spans tokenized
correctly. Comparing token identifiers on natural Portuguese sentences, the
converted tokenizer agreed with the training SentencePiece on zero of five probe
strings, all failures occurring on capitalized words and proper nouns.

The effect is severe and invisible in aggregate perplexity over long windows, which
is exactly why it is dangerous. Evaluated with the uncorrected tokenizer,
Manac\'a-1B scored 25.0 on LAMBADA-PT with a token-level perplexity near $10^{6}$,
which would read as a weak model. The fix is to inject the missing normalizer at the
JSON level, setting it to the sequence NFKC then Lowercase, which reproduces the
training tokenization on all five probe strings. With the corrected tokenizer the
same LAMBADA-PT accuracy rose to 45.3 and the perplexity fell to 17.3. We report all
downstream numbers with the corrected tokenizer, release it for reuse, and note the
general lesson: when a model is trained under nontrivial tokenizer normalization, the
evaluation tokenizer must be verified against the training tokenizer before any score
is trusted.

\section{Model Export and Conversion}
\label{sec:convert}

Evaluating with community harnesses requires the model in the HuggingFace format, and
the conversion from the Megatron checkpoint is not mechanical for this architecture.
The public saver shipped with the LLM-jp fork targets standard Llama-style models and
makes two assumptions that do not hold here, so we wrote a converter that corrects
both. The first is bias. Manac\'a-1B carries learned biases in every linear layer, and
in the attention projections their magnitude is not negligible; a converter that
silently drops them, as the stock saver does, produces a model that loads without
error and generates fluent-looking text while being subtly wrong. Our converter copies
the query, key, value, and output biases and both feed-forward biases, sets the
corresponding \texttt{attention\_bias} and \texttt{mlp\_bias} flags on the target
config, and aborts with an explicit message if the installed \texttt{transformers} is
too old to represent them, so a version mismatch can never degrade into a silent
partial export.

The second complication is the tensor layout. Megatron fuses the query, key, and value
projections into one packed matrix and, under grouped-query attention, interleaves the
eight key-value groups with their query heads; the converter unpacks each group into
the separate query, key, and value matrices that the HuggingFace model expects. The
SwiGLU feed-forward is stored as a single fused matrix whose two halves are the gate
and up projections, and these are split at the feed-forward width. Because the model is
trained with untied input and output embeddings, the output projection is copied to a
separate \texttt{lm\_head} rather than shared with the token embeddings, and the fused
RMSNorm weights that Megatron keeps inside the attention and feed-forward blocks are
mapped to their standalone HuggingFace counterparts. The result is a standard
\texttt{LlamaForCausalLM} in bfloat16 safetensors with a padded vocabulary of 64{,}128.

We verify the conversion at three levels rather than trusting it. First, the state
dictionary is loaded with an exhaustive key check that aborts on any missing or
unexpected tensor, so an incomplete or misdirected mapping cannot pass. Second, the
reloaded model greedy-generates short continuations for a handful of Portuguese prompts,
a qualitative check that the attention splits and biases are coherent enough to produce
grammatical text. Third, and most important for evaluation, the exported tokenizer is
checked identifier by identifier against the training SentencePiece on probe strings
that deliberately include capitalization, which is the check that surfaced the
normalizer bug of Section~\ref{sec:tok}. We note honestly what this does not include:
there is no numerical comparison of Megatron and HuggingFace logits on identical
inputs, so the guarantee is exact key coverage and exact tokenization rather than
bit-level output equivalence.

\section{Evaluation Protocol}

We evaluate every model on the same four benchmarks, with one harness and one set of
prompts, and we report uncertainty for every number. CALAME-PT \cite{gloria} measures
last-word prediction on 2{,}075 native Portuguese passages and is scored by greedy
generation of the final word. ARC-Challenge-PT and HellaSwag-PT are the Portuguese
splits of the Okapi \cite{okapi} machine translations of ARC \cite{arc} and HellaSwag
\cite{hellaswag}, scored by length-normalized log-likelihood at 25-shot and 10-shot
respectively, matching the protocol reported by Tucano. LAMBADA-PT \cite{lambada} is
the Portuguese translation used by Tucano, scored by log-likelihood at zero shot. The
multiple-choice and log-likelihood tasks run through the EleutherAI
lm-evaluation-harness \cite{lmeval}; CALAME-PT runs through our own scorer so that
Manac\'a-1B can use its training SentencePiece directly.

The baselines span the open Portuguese and multilingual landscape at comparable
scale: TeenyTinyLlama-160m and 460m \cite{ttl}, Tucano-160m, 630m, 1b1, and 2b4
\cite{tucano}, Gl\'orIA-1b3 \cite{gloria}, mGPT-1b3 \cite{mgpt}, and Sabi\'a-7B
\cite{sabia}. Uncertainty is reported as the standard error of each metric (binomial
for the accuracies, the harness estimate for the log-likelihood tasks). To compare
two models we use the paired McNemar test and a paired bootstrap over the shared
examples, rather than the overlap of marginal confidence intervals, because the two
models see identical items and their errors are correlated.

We validate the harness against previously published numbers before trusting any of our
own, and Table~\ref{tab:valid} reports the check. On CALAME-PT the four Tucano models,
which share the SentencePiece tokenizer family with Manac\'a-1B, reproduce their
published accuracies within about one point. On the log-likelihood tasks Tucano-1b1
reproduces within about one to one and a half points on the two multiple-choice
benchmarks and within about three points on LAMBADA-PT, the larger gap there reflecting
that the published LAMBADA-PT number uses a generation-based protocol while ours uses
log-likelihood. Two baselines do not reproduce cleanly, and we flag this rather than
hide it: the byte-level-BPE models Gl\'orIA-1b3 and mGPT-1b3 score about seven to eight
points above their published CALAME-PT values under our generation-based scorer, because
byte-level continuation tokenization favors the last-word match in that protocol. We
therefore treat the CALAME-PT numbers for those two models as provisionally inflated,
which if anything strengthens rather than weakens the comparisons in which Manac\'a-1B
already leads or ties them.

\begin{table}[t]
\centering
\caption{Harness validation against published numbers. CALAME-PT rows compare our
scorer to the published Tucano and baseline values; the Tucano-1b1 rows compare our
lm-evaluation-harness run to published log-likelihood numbers. The SentencePiece models
reproduce within about one point; the two byte-level-BPE models (marked) are inflated by
our generation-based CALAME-PT protocol.}
\label{tab:valid}
\small
\begin{tabular}{llrrr}
\toprule
Model & Benchmark & Ours & Published & $\Delta$ \\
\midrule
Tucano-160m  & CALAME-PT     & 52.53 & 52.31 & $+0.22$ \\
Tucano-630m  & CALAME-PT     & 56.63 & 56.55 & $+0.08$ \\
Tucano-1b1   & CALAME-PT     & 59.08 & 58.24 & $+0.84$ \\
Tucano-2b4   & CALAME-PT     & 59.57 & 59.06 & $+0.51$ \\
Gl\'orIA-1b3$^{\dagger}$ & CALAME-PT & 60.39 & 52.79 & $+7.60$ \\
mGPT-1b3$^{\dagger}$     & CALAME-PT & 55.57 & 47.14 & $+8.43$ \\
\midrule
Tucano-1b1   & ARC-Ch-PT     & 29.66 & 30.43 & $-0.77$ \\
Tucano-1b1   & HellaSwag-PT  & 44.23 & 42.84 & $+1.39$ \\
Tucano-1b1   & LAMBADA-PT    & 31.50 & 34.70 & $-3.20$ \\
\bottomrule
\end{tabular}
\\[2pt]
{\footnotesize $^{\dagger}$ byte-level-BPE tokenizer; over-scored by the
generation-based CALAME-PT protocol.}
\end{table}

\section{Results and Discussion}

\begin{table}[t]
\centering
\caption{Accuracy (\%) with standard error on four Brazilian-Portuguese benchmarks,
all models under one harness. CALAME-PT is scored by greedy last-word generation;
ARC-Challenge-PT (25-shot), HellaSwag-PT (10-shot), and LAMBADA-PT (0-shot) by
log-likelihood. Manac\'a-1B uses its training tokenizer. Best per column in bold.}
\label{tab:main}
\small
\begin{tabular}{lrcccc}
\toprule
Model & Params (B) & CALAME-PT & ARC-Ch-PT & HellaSwag-PT & LAMBADA-PT \\
\midrule
TTL-160m       & 0.16 & 47.33 \tiny$\pm$1.10 & 25.38 \tiny$\pm$1.27 & 29.73 \tiny$\pm$0.48 & 21.21 \tiny$\pm$0.57 \\
Tucano-160m    & 0.16 & 52.53 \tiny$\pm$1.10 & 25.04 \tiny$\pm$1.27 & 33.56 \tiny$\pm$0.49 & 25.62 \tiny$\pm$0.61 \\
TTL-460m       & 0.46 & 51.23 \tiny$\pm$1.10 & 27.09 \tiny$\pm$1.30 & 34.47 \tiny$\pm$0.49 & 22.18 \tiny$\pm$0.58 \\
Tucano-630m    & 0.63 & 56.63 \tiny$\pm$1.09 & 27.52 \tiny$\pm$1.31 & 39.96 \tiny$\pm$0.51 & 31.03 \tiny$\pm$0.64 \\
Tucano-1b1     & 1.10 & 59.08 \tiny$\pm$1.08 & 29.66 \tiny$\pm$1.34 & \textbf{44.23} \tiny$\pm$0.52 & 31.50 \tiny$\pm$0.65 \\
Gl\'orIA-1b3   & 1.30 & 60.39 \tiny$\pm$1.07 & 24.44 \tiny$\pm$1.26 & 25.83 \tiny$\pm$0.46 & 35.30 \tiny$\pm$0.67 \\
mGPT-1b3       & 1.30 & 55.57 \tiny$\pm$1.09 & 23.93 \tiny$\pm$1.25 & 25.42 \tiny$\pm$0.45 & 37.38 \tiny$\pm$0.67 \\
\textbf{Manac\'a-1B} & 1.72 & \textbf{60.63} \tiny$\pm$1.07 & 27.18 \tiny$\pm$1.30 & 41.61 \tiny$\pm$0.51 & \textbf{45.31} \tiny$\pm$0.69 \\
Tucano-2b4     & 2.40 & 59.57 \tiny$\pm$1.08 & 30.85 \tiny$\pm$1.35 & 48.63 \tiny$\pm$0.52 & 34.35 \tiny$\pm$0.66 \\
Sabi\'a-7B     & 7.00 & \textbf{63.23} \tiny$\pm$1.06 & \textbf{46.67} \tiny$\pm$1.46 & \textbf{64.55} \tiny$\pm$0.50 & \textbf{63.67} \tiny$\pm$0.67 \\
\bottomrule
\end{tabular}
\end{table}

The strongest result is that Manac\'a-1B leads every model below the 7B scale on
last-word prediction, and the paired tests make this precise. On LAMBADA-PT
Manac\'a-1B reaches 45.31, above Tucano-1b1 (31.50), Tucano-2b4 (34.35),
Gl\'orIA-1b3 (35.30), and mGPT-1b3 (37.38); the paired McNemar test rejects equality
against all of them at $p<10^{-4}$, and only Sabi\'a-7B, a four-times-larger model,
scores higher. The pattern repeats on CALAME-PT, where Manac\'a-1B is the highest of
the sub-7B group at 60.63, although here the margins over Tucano-1b1 and Tucano-2b4
are within noise (paired McNemar $p=0.13$ and $p=0.29$). Last-word prediction rewards
fluent modeling of Portuguese morphology, and it is where the native 64k tokenizer
and the from-scratch Portuguese training pay off most clearly.

The picture inverts on the tasks that reward broad training exposure. On HellaSwag-PT
Manac\'a-1B (41.61) is significantly below Tucano-1b1 (44.23) and Tucano-2b4 (48.63),
both $p<10^{-4}$, yet significantly above the same-size Gl\'orIA-1b3 (25.83) and
mGPT-1b3 (25.42). On ARC-Challenge-PT all sub-2B models sit near the 25 percent chance
line, and no pairwise difference among them is significant; Manac\'a-1B is
indistinguishable from Tucano-1b1 (paired $p=0.059$) and falls significantly below
only Tucano-2b4 ($p=0.0047$) and Sabi\'a-7B. Two readings follow. First, the Tucano
models, trained on more than an order of magnitude more tokens, convert that exposure
into commonsense and reasoning accuracy that Manac\'a-1B, trained to a
near-compute-optimal budget, does not match at equal parameter count. Second, at the
same parameter budget Manac\'a-1B clearly outperforms
the European-Portuguese and multilingual baselines, which is the comparison that
isolates the value of dedicated Brazilian-Portuguese pretraining.

The near-chance ARC-Challenge-PT result deserves a direct explanation, because it holds
for every sub-2B model in Table~\ref{tab:main} and not only for Manac\'a-1B.
ARC-Challenge is a deliberately hard, four-way multiple-choice science-reasoning task,
and answering it requires discriminating among options whose surface forms are similar,
a capability that base language models acquire late, with scale and with instruction
tuning, rather than from language modeling alone. The only models that clear chance
here are the ones with far more capacity or far more training: Sabi\'a-7B reaches 46.67
and Tucano-2b4 edges to 30.85, while everything from 0.16B to 1.72B stays inside a band
around 25 percent whose internal differences are not significant. Two factors specific
to this benchmark compound the effect. It is a machine translation of the English ARC,
so part of its scientific vocabulary and its answer distractors are translated rather
than native, and length-normalized log-likelihood over four short options is a blunt
signal at this scale. The honest reading is that near-chance ARC-Challenge-PT is the
expected behavior of a 1B-scale Brazilian-Portuguese base model, not a defect of this
particular one, and it is consistent with the published Tucano numbers we reproduce.
It is also the clearest indication of where instruction tuning and additional scale,
both left to future work, would help.

Table~\ref{tab:paired} reports the paired comparisons in full on the three
log-likelihood benchmarks, where per-example predictions are available. Each entry is
the signed accuracy difference of Manac\'a-1B minus the baseline in points, with a star
for a paired McNemar rejection at $p<0.05$. The table makes the structure of the
results explicit: on LAMBADA-PT every difference is significant, positive against all
sub-7B baselines and negative only against Sabi\'a-7B; on HellaSwag-PT the sign tracks
training exposure, positive against the smaller and non-Brazilian models and negative
against Tucano-1b1 and Tucano-2b4; and on ARC-Challenge-PT only the two largest and
most heavily trained baselines separate from Manac\'a-1B at all.

\begin{table}[t]
\centering
\caption{Paired comparison of Manac\'a-1B against each baseline on the three
log-likelihood benchmarks. Entries are the accuracy difference (Manac\'a-1B minus
baseline) in percentage points; $^{\ast}$ marks a paired McNemar test significant at
$p<0.05$ over the shared examples. Positive favors Manac\'a-1B.}
\label{tab:paired}
\small
\begin{tabular}{lrrr}
\toprule
Baseline & LAMBADA-PT & HellaSwag-PT & ARC-Ch-PT \\
\midrule
TTL-160m     & $+24.10^{\ast}$ & $+11.89^{\ast}$ & $+1.79$ \\
Tucano-160m  & $+19.70^{\ast}$ & $+8.05^{\ast}$  & $+2.14$ \\
TTL-460m     & $+23.13^{\ast}$ & $+7.14^{\ast}$  & $+0.09$ \\
Tucano-630m  & $+14.28^{\ast}$ & $+1.65^{\ast}$  & $-0.26$ \\
Tucano-1b1   & $+13.82^{\ast}$ & $-2.62^{\ast}$  & $-2.39$ \\
Gl\'orIA-1b3 & $+10.01^{\ast}$ & $+15.78^{\ast}$ & $+2.56$ \\
mGPT-1b3     & $+7.94^{\ast}$  & $+16.33^{\ast}$ & $+2.82$ \\
Tucano-2b4   & $+10.96^{\ast}$ & $-7.02^{\ast}$  & $-3.68^{\ast}$ \\
Sabi\'a-7B   & $-18.36^{\ast}$ & $-22.94^{\ast}$ & $-19.49^{\ast}$ \\
\bottomrule
\end{tabular}
\end{table}

\begin{figure}[t]
\centering
\includegraphics[width=\textwidth]{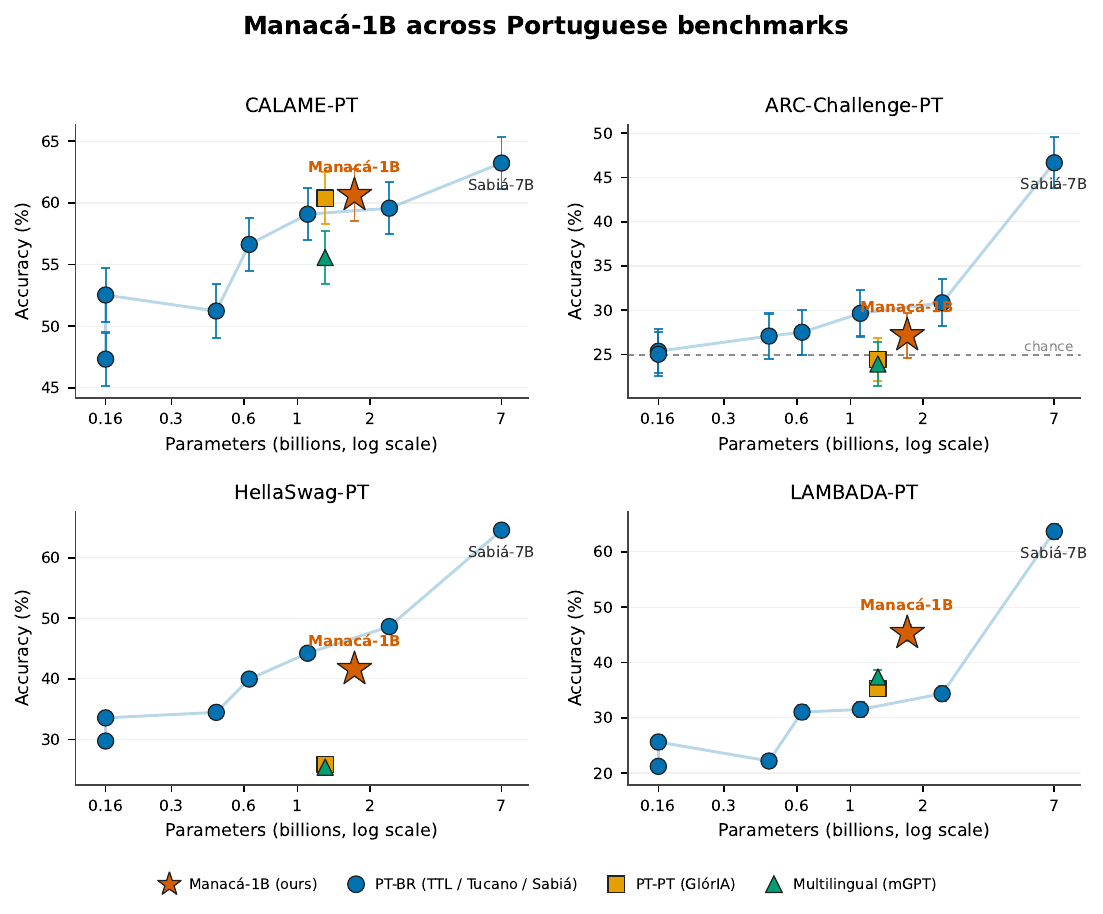}
\caption{Manac\'a-1B across four Brazilian-Portuguese benchmarks. Accuracy versus
parameter count (log scale) for Manac\'a-1B and nine open baselines under one harness;
error bars are 95\% confidence intervals. Marker shape encodes model family, so the
figure remains readable in grayscale and under color-vision deficiency \cite{wong}.
Pairwise
significance is reported in the text and in the released tables.}
\label{fig:panel}
\end{figure}

Figure~\ref{fig:panel} places these numbers on a common scale. Across the four panels
Manac\'a-1B sits on or above the Brazilian-Portuguese size trend on the two last-word
tasks, slightly below it on HellaSwag-PT, and inside the near-chance cluster on
ARC-Challenge-PT. The single sharpening that the paired test provides over a marginal
comparison is on ARC-Challenge-PT against Tucano-2b4, which moves from borderline
under the unpaired two-proportion test ($p=0.05$) to significant under McNemar
($p=0.0047$); all other verdicts are unchanged. Read together, the honest summary is
parity with the best open Brazilian-Portuguese models of comparable size on language
modeling, a clear advantage over same-size open peers from other varieties and
languages, and a reasoning gap to the far more data-rich models that is consistent
with the training budget.

\section{Limitations}

The scope of this release is deliberately narrow, and we state its boundaries
plainly. Manac\'a-1B is a base model trained on about 42 billion update tokens, close to
the compute-optimal ratio but far below the token counts of its strongest baselines, and
those update tokens are about two epochs over a curated 20-billion-token corpus rather
than a single pass over a larger one; its weaker reasoning scores should be read as a
consequence of that budget difference and not as an architectural ceiling.
The evaluation covers four benchmarks that emphasize last-word prediction and
translated multiple-choice tasks, and it does not yet include the native Brazilian
exam and inference tasks of the Open Portuguese LLM Leaderboard, which we consider
essential for full comparability with the Portuguese literature. LAMBADA-PT is scored
here by log-likelihood, which approximates but does not exactly reproduce the
generation-based protocol of its original release, and CALAME-PT is scored by
generation while the other three are scored by log-likelihood, so the four columns of
Table~\ref{tab:main} should be compared within a column rather than across. Finally,
we release only a base model; no instruction-tuned variant is evaluated here.

\section{Future Work}

The natural next steps extend the evaluation and the training rather than the
architecture. We will add the native Brazilian benchmarks of the Open Portuguese LLM
Leaderboard, evaluated on the base model under the same paired protocol, and general
multilingual baselines such as Llama-3.2 and Qwen2.5 for external positioning. A
controlled ablation of a Brazilian versus European Portuguese variant filter, whose
data and training scripts are already prepared in the repository, will test whether
variety-specific filtering improves downstream accuracy at fixed compute. We will also
extend the token budget well beyond the compute-optimal point, into the over-trained
regime that the strongest baselines occupy, to separate the effect of extra data from
that of scale at a fixed parameter count, and we will train and evaluate an
instruction-tuned Manac\'a for the tasks where following instructions is the point.

\section{Reproducibility and Availability}

Every number in this paper can be recomputed from the public repository. It contains
the containerized training, conversion, and evaluation pipeline, the raw training log,
the raw per-model evaluation logs, the per-example prediction vectors used for the
paired tests, the scripts that produce the tables and Figure~\ref{fig:panel}, and the
corrected tokenizer described in Section~\ref{sec:tok}. The evaluation harness version
is pinned, the benchmark datasets are named by their public identifiers, and the
statistical tests are implemented as small, auditable scripts rather than computed by
hand. The code is released at
\url{https://github.com/Instituto-IA-LNCC/manaca-1b-base}, and the model weights,
together with the corrected tokenizer of Section~\ref{sec:tok}, are published under a
CC BY 4.0 license at \url{https://huggingface.co/menezesbruno/manaca-1b-base}.

\section{Conclusion}

Manac\'a-1B shows that a small, honestly evaluated, fully reproducible model is a
useful contribution to Brazilian-Portuguese natural language processing even without
record-setting scale. Measured with paired significance under one harness, the model
matches the best open Portuguese models of its size on language modeling and clearly
surpasses same-size peers from other varieties, while an explicit token budget
accounts for its reasoning gap to far larger corpora. The tokenizer-fidelity finding
generalizes beyond this model: whenever a model is trained under nontrivial tokenizer
normalization, the evaluation tokenizer must be verified against the training
tokenizer, or otherwise-careful comparisons can be silently wrong. We hope that the
released model, corrected tokenizer, and reproducible protocol make it easier for the
community to build on this work and to hold it, and future Portuguese models, to a
measurable standard.

\section*{Acknowledgments}

This work was carried out at the Artificial Intelligence Institute of the National
Laboratory for Scientific Computing (LNCC), Brazil, an institution of the Brazilian
Ministry of Science, Technology and Innovation, as part of an ongoing cooperation
between LNCC and the National Institute of Informatics (NII), Japan, under a
Memorandum of Understanding between the two institutions. We gratefully acknowledge
the LLM-jp project, the open large-language-model initiative organized by NII, whose
openly released methodology, tooling, and models made this work possible: Manac\'a-1B
follows the LLM-jp-3.1-1.8B architecture and training recipe, is trained with the
LLM-jp fork of Megatron-LM, and its evaluation design is informed by the LLM-jp
evaluation practice. We thank our colleagues at LNCC and at NII/LLM-jp for the
technical exchange that shaped this effort, and LNCC for the computational
infrastructure on which the model was trained. Any remaining errors are our own.

\section*{Declaration on the Use of AI Tools}

In the interest of transparency and in line with current publishing guidance on the
use of generative AI in scientific writing, we disclose that an AI assistant (Claude,
developed by Anthropic) was used solely to support the review of this manuscript: to
check the text for clarity, consistency, and structure, and to help verify the
references. Consistent with COPE and ICMJE guidance, the AI tool is not and cannot be
an author, and it bears no responsibility for the work. The authors conceived and
designed the study, carried out all of the experiments, wrote the manuscript, produced
and verified every dataset, number, table, and figure, checked each reference against
its primary source, and take full responsibility for the content and for any remaining
errors. The work is original: no text or results were plagiarized, all external sources
are cited, and every scientific claim reflects the authors' own analysis and judgment.

\end{document}